\documentclass[10pt,twocolumn,letterpaper]{article}

\usepackage{iccv}              % To produce the CAMERA-READY version

\usepackage{graphicx}
\usepackage{caption}
\usepackage{colortbl}
\usepackage[subtle]{savetrees}
\usepackage{hyperref}
\usepackage{makecell}
\usepackage[accsupp]{axessibility}
\usepackage{multirow}
\definecolor{iccvblue}{rgb}{0.21,0.49,0.74}
\PassOptionsToPackage{pagebackref,breaklinks,colorlinks,allcolors=iccvblue}{hyperref}

\def\confName{ICCV}
\def\confYear{2025}
\title{Unified Supervision For Vision-Language Modeling in 3D Computed Tomography}

\author{
\begin{tabular}{@{}c@{}}
    Hao-Chih Lee$^1$ \qquad
    Zelong Liu$^1$ \qquad
    Hamza Ahmed$^1$ \qquad 
    Spencer Kim$^1$ \qquad 
    Sean Huver$^2$ \qquad  \\
     Vishwesh Nath$^2$ \qquad 
    Zahi A. Fayad$^1$ \qquad
Timothy Deyer$^{3,4}$ \qquad
 Xueyan Mei$^1$ \qquad
\end{tabular} \\ 
\begin{tabular}{@{}c@{}}
    $^1$BioMedical Engineering and Imaging Institute, Icahn School of Medicine at Mount Sinai \\
    $^2$ NVIDIA \quad
    $^3$East River Medical Imaging \quad 
    $^4$Department of Radiology, Cornell Medicine \qquad 
\end{tabular}
}
\begin{document}
\maketitle
\begin{abstract}
General-purpose vision-language models (VLMs) have emerged as promising tools in radiology, offering zero-shot capabilities that mitigate the need for large labeled datasets. However, in high-stakes domains like diagnostic radiology, these models often lack the discriminative precision required for reliable clinical use. This challenge is compounded by the scarcity and heterogeneity of publicly available volumetric CT datasets, which vary widely in annotation formats and granularity. To address these limitations, we introduce Uniferum, a volumetric VLM that unifies diverse supervision signals, encoded in classification labels and segmentation masks, into a single training framework. By harmonizing three public 3D CT datasets with distinct annotations, Uniferum achieves state-of-the-art performance, improving AUROC on the CT-RATE benchmark by 7\% compared to CLIP-based and conventional multi-label convolutional models. The model demonstrates robust out-of-distribution generalization, with observed evidence of unexpected zero-shot performance on the RAD-CHEST and INSPECT datasets. Our results highlight the effectiveness of integrating heterogeneous annotations and body segmentation to enhance model performance, setting a new direction for clinically reliable, data-efficient VLMs in 3D medical imaging.
\end{abstract}    
\section{Introduction}
\label{sec:intro}
\begin{figure}[t!]
    \centering
    \includegraphics[width=\columnwidth]{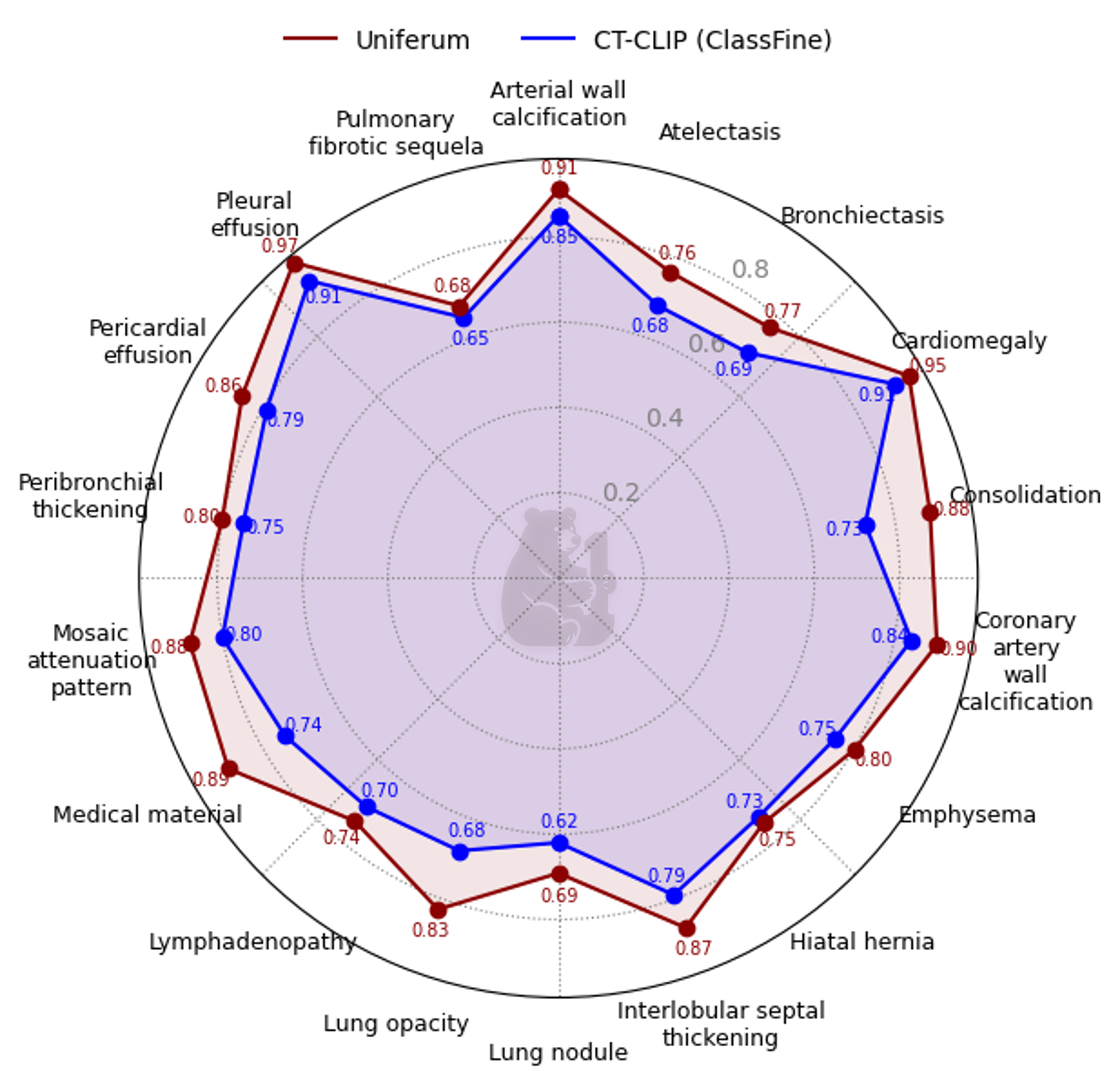}
    \caption{Performance comparison on CT-RATE. We compared Uniferum and CT-CLIP with ClassFinetuning on CT-RATE testing set. Both models were trained solely on CT-RATE.}
    \label{fig:fig1}
\end{figure}
General-purpose vision-language models (VLMs) have recently attracted widespread attention for their potential to assist in radiology-based diagnosis. VLMs leverage the scale and rich information embedded in paired medical scans and radiology reports. These models demonstrate zero-shot capabilities, the ability that can solve unseen tasks without requiring any labeled data, that can address key limitations of traditional supervised approaches in medical imaging. In particular, contrastive language-image pre-training (CLIP) \cite{radford2021learning}, after being trained on millions of image–text pairs, can be applied “out of the box” to new classification tasks by matching image embeddings to text prompts. This zero-shot approach has sparked interest in medical imaging research, where labeled data is both scarce and costly to obtain. 

However, in a high-risk, highly regulated environment such as diagnostic radiology, coarse zero-shot capability alone is insufficient. Clinical decision support tools must meet stringent performance requirements as small differences in sensitivity or specificity can translate into significant patient safety implications. Although CLIP-like models offer broad generalization, their zero-shot outputs frequently lack the fine-tuned discriminative power needed to reliably detect subtle pathologies or to meet predefined performance thresholds in a clinical workflow \cite{zhang2024mediclip,woerner2024navigating}.

How do we go about developing a model of high discriminative power given that volumetric CT images resources, unlike hundreds of millions of natural images available on the Web, are scarce and fragmented? Public volumetric CT datasets rarely exceed tens of thousands of scans. In addition, each dataset has its own styles and choices of annotations. Some datasets provide structured radiology reports, others supply disease labels of varying granularity, and still others offer segmentation masks for specific organs or lesions. For instance, one dataset might include 18 broad pathological labels, while another lists over 80 more detailed categories of medical findings \cite{hamamci2024developing,draelos2021machine}; many benchmarks focus on segmentation tasks that do not directly translate into classification objectives \cite{wasserthal2023totalsegmentator, murphy2009large}. The challenge lies in harmonizing all these resources (different datasets with different annotations) in assembling a large scale data set for model development.

To address these challenges, we propose a vision-language model, dubbed Uniferum, that unifies various training signals within one framework. We collect and harmonize three public volumetric CT datasets, each with different annotation labels, and design a training strategy that accommodates multiple supervision modalities.  Our model shows superior performance on CT-RATE compared with CLIP based model and standard multi-label volumetric convolutional neural networks. Additionally Uniferum demonstrates surprising zero-shot classification performance on completely with-held datasets. We observed that performance increases with the number of supervised learning tasks. By unifying heterogeneous annotations into a single VLM framework, Uniferum addresses both fragmented data resources and the need for high discriminative power in 3D CT interpretation. This approach paves the way toward clinically reliable, data-efficient vision-language models for radiology.

Our key contributions include
\begin{enumerate}
\item We present a framework that unifies diverse supervision signal, including classification labels and segmentation masks, from heterogeneous 3D CT datasets.
\item The proposed volumetric vision-language model that improves the state-of-the-art on CT-RATE by 7\% in AUROC. We thoroughly evaluate its performance, both out-of-distribution and zero-shot, on two additional large-scale volumetric CT datasets: RAD-CHEST and INSPECT.
\item Our results demonstrate that the proposed model effectively integrates diverse training signals from both class labels and segmentation masks. Notably, we find that incorporating body segmentation provides a universal strategy for enhancing performance.
\end{enumerate}
The implementation is publicly available at \href{https://github.com/howchihlee/uniferum}{https://github.com/howchihlee/uniferum}.

\section{Related Work}
\label{sec:formatting}

\textbf{VLM in General Domain:} 
\vspace{\baselineskip}
\noindent \textbf{2D VLM in Radiology} 

Vision–language models (VLMs) have made significant advances in biomedical imaging, particularly in interpreting two dimensional images such as chest x-ray images. Early efforts, such as CheXpert \cite{irvin2019chexpert}, demonstrated the feasibility of mapping radiology reports to multi-label pathology classification. RadGraph \cite{jain2021radgraph} extended the idea by enhancing isolated labels with a structured knowledge graph, laying the foundation for further evaluation \cite{yu2023evaluating} and knowledge-enhanced modeling \cite{zhang2023knowledge}.
Contrastive learning has gained considerable attention in 2D vision–language modeling. BiomedCLIP \cite{zhang2023biomedclip} was pre-trained on a wide range of biomedical imaging tasks using a dataset of 15 million biomedical image–text pairs derived from 4.4 million scientific articles. GLoRIA \cite{huang2021gloria} and subsequent models improved upon CLIP by incorporating fine-grained information specific to anatomical regions. MedCLIP \cite{wang2022medclip} introduced semantic matching loss that utilizes entity recognition to estimate semantic similarity between text pairs to mitigate the data hungry issue of CLIP for medical imaging. Particularly relevant to our work, Xu et al. \cite{xu2023learning} proposed an encoder–decoder multi-task transformer to tackle tasks such as classification, segmentation, and localization. Their model, however, was primarily designed for generative tasks rather than optimized for discriminative performance.

\vspace{\baselineskip}
\noindent \textbf{3D volumetric image analysis and vision language modeling}
Draelos et al. \cite{draelos2021machine} developed an image-based convolutional neural network on RADCHEST to classify 83 abnormalities. While varying accuracy depending on categories, Draelos et al. demonstrated that performance improved as the number of labels increased. Huang et al. \cite{huang2023inspect} combined ResNetV2 and transformer layers to classify plumary emblosim patients and the associated prognostic risks. They found that models integrating both EHR and CT images performed better than single-modality models, while the image-based model performed the worst. Joint classification and segmentation \cite{lu2025automatic, he2023joint, mehta2018net} were investigated in the multi-task learning setting and found to perform better when integrating both tasks. Unlike our model, these models rely on paired classification and segmentation masks. Li et al. \cite{li2025towards} investigated text-driven segmentation for open-ended segmentation but they did not pursue tasks beyond segmentation. 

Hamamci developed generalist models from a multimodal Dataset for 3D CT chest scans and demonstrated clip’s zero-shot classification performs better than supervised convolutional neural networks \cite{hamamci2024developing}. Merlin \cite{blankemeier2024merlin}, a clip based model, was recently developed based on 15,331 3D CT scans, primarily focused on the abdomen region. There are many other VLM that demonstrate generalist capability but the discriminative power on detecting abnormality from 3D CT scans have not been thoroughly examined \cite{lai2024e3d, bai2024m3d,chen20243d,wu2023towards,yu2025umit}.

\raggedbottom
\section{Method}
\subsection{Task format and notations}
In this work, we propose a unified model for volumetric medical image analysis that unifies classification and segmentation labels by conditioning on a task-specific natural language description. Let ($X$, $\cdot$) denote a volumetric medical image $X\in \mathbb{R}^{n\times m\times s}$ and a supervision labels. We assume that the image volume has either a corresponding binary classification label $y\in \{0,\; 1\}$ or a paired binary segmentation mask $M\in \{0,1\}^{n \times m \times s}$, where each voxel indicates whether it belongs to a structure of interest. 

Each input pair is associated with a task description $t$, which defines the clinical or anatomical objective. For classification, a typical $t$ might be: “Diagnose the presence of coronary artery disease in the heart,” where y indicates presence (1) or absence (0) of the condition. For segmentation, $t$ could be: “Segment the lungs in the image,” where $M$ highlight the regions corresponding to the lungs in the CT scan $X$. We refer to the tuple ($X$, $y$ or $M$, $t$) as a vision language task in the following text. Note that a single image volume can be associated with multiple labels $y$ and masks $M$ in this context. For example, an image volume with multiple classification labels can be decomposed into several distinct classification tasks, all sharing the same input $X$. See Section 3.3 for details on how such tasks are constructed from a dataset. 

Notably, we do not assume that labels are paired. Neither joint label–mask pairs nor multi-label annotations are required. In multi-label classification, most models estimate a function of the form $f(x)=\mathbf{y} \in \mathbb{R}^k$, where $
\mathbf{y}=(y_1,\;y_2,\;\ldots,\; y_k)$ is a multi-label output vector. In contrast, we reformulate the problem as estimating a function conditioned on a task-specific description $t_i$, i.e., $f(x, t_i) = y_i$. This formulation decouples individual label predictions, allowing each sample to be reused across multiple tasks. As a result, the effective training set size increases from $n$ to $n\times k$, and opens up the door to maximizing the utility of fragmented labeled data.

\subsection{Model}
Our model is an encoder-only VLM that comprises a vision encoder $f_v$ and a transformer $f_t$ that integrates two distinct embedding vectors: one for chest CT volumes and another for task description, respectively (Figure 2). We use $E_v = f_v(X) \in R^{m/d \times n/d \times s/d}$ where $d$ is the overall factor that downsample the input tensor through the forward propagation. The flattened output of the vision encoder $E_v$ is concatenated with the token embeddings of the task description $E_t$, along with two additional embedding vectors: a classification token $v_{\textup{cls}}$ and a separator token $v_{\textup{sep}}$. Positional embeddings are added to this sequence before it is passed through a stack of bidirectional transformer layers. The two additional embedding vectors serve as the CLS token for classification and SEP token to separate vision and text embedding respectively. 
The final output is $E = \textup{TransformerLayers}(Z)$
where \[Z = \textup{concat}([v_\textup{cls},\;E_v,\;v_\textup{sep},\;E_t]) + E_\textup{pos}.\]
We split $E$ into a vector and a sequence of embeddings. The first is the transformed CLS embedding vector $E_\textup{cls}$ which is positioned at the first entry of $E$. The following $r = mns/d^3$ vectors in $E$ form a sequence denoted as $E_s \in \mathbb{R}^{r \times f_h}$, which serves as embeddings to predict segmentation mask. $f_h$ is the dimension of hidden space.

For classification tasks, we passed $E_\textup{cls}$  into a linear layer to transform it into a prediction logit, where binary cross-entropy is used for model training.
For segmentation tasks, we apply a linear layer to the segmentation embeddings $E_s$ to predict a binary mask. While this method is straightforward, it must address two main challenges. First, predicting full-resolution 3D volumetric images is memory-intensive. Second, the 3D convolutional encoder reduces spatial resolution by a factor of $d$ (the downsampling factor). To balance these constraints, we adopt a linear layer that up-samples the segmentation embeddings $E_s$ to an intermediate resolution.

Specifically, we first apply a max pooling operation with filter size $d/u$ to the ground truth binary mask. We then divide the volume into non-overlapping patches of size $u \times u \times u$, flatten each patch, and stack them into a binary target array of shape $(mns/d^3, u^3)$. Accordingly, the linear layer is designed to map each embedding in $E_s$ to a vector of size $u^3$, producing an output that aligns with the target patch-wise representation. Although our goal is not to produce fine-grained segmentation, but rather to integrate fine-grained localization information into the classification model, the final output can still be unpacked for visualization purposes.

During model training, we mixed classification and segmentation in one batch and calculated loss only for the task whose target label is available. Binary cross-entropy was used for classification tasks, while focal loss \cite{lin2017focal} was applied to measure the discrepancy between predicted logits and the target masks in segmentation tasks. To address the issue that focal loss values can diminish rapidly during training, we scaled the focal loss by a factor of 10.

\begin{figure}[t!]
    \centering
    \includegraphics[width=\columnwidth]{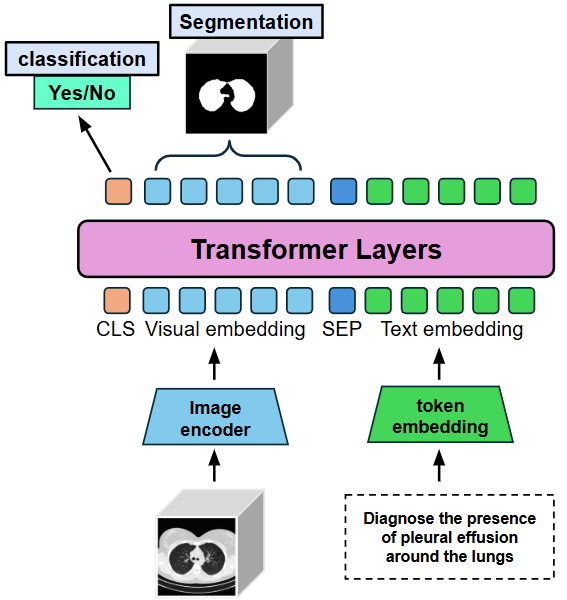}
    \caption{model architecture of Uniferum.}
    \label{fig:fig2}
\end{figure}

\subsection{Creating training data}
Designing and creating training data is a critical part of our work. For each label in the dataset we create a task description. We categorize tasks in the following three groups: 
\begin{enumerate}
\item For diagnostic tasks we used the template “Diagnose the presence of [label name] in/around the [organ]” to generate the task description. For example, for the label of pleural effusion, the task description associated with it is “Diagnose the presence of pleural effusion around the lungs”. We reviewed and reformatted the labels into readable text if necessary. 
\item For prognostic tasks, we used the template “Predict the risk of  [label name] in [x months]”. For example, “Predict the risk of mortality in 6 months” describes the 6-month mortality label in INSPECT.
\item For segmentation tasks, we applies the template “Segment [label name] in the image.” We generate 60 organ segmentation tasks that include 53 labels predicted by Total Segmentator and 7 derived tasks by combining TotalSegmentator labels into general categories such as “ribs” and “lungs”. We include an additional nodule segmentation task using 601 masks from the LUNA16\cite{murphy2009large} dataset. 
\end{enumerate}
To handle the issue of sample imbalance, we randomly sampled negative examples to match the number of positive examples for each diagnostic and prognostic task. Additionally, we randomly sampled an extra 10\% samples from the dataset to assign an organ segmentation task. We upsampled the LUNA16 samples tenfold to a total of 6,010 samples as an add-on for datasets augmented with segmentation tasks. 

\subsection{Datasets}
We utilized three publicly available 3D CT datasets, including CT-RATE, RAD-CHEST, and INSPECT, for model development and benchmarking. Each dataset provides volumetric imaging data along with task-specific annotations. Below, we summarize the key statistics of three dataset in table 1 and describe the datasets in detail below:
\subsubsection{CT-RATE}
CT-RATE \cite{hamamci2024developing} is a large-scale dataset of 50,188 non-contrast 3D chest CT volumes paired with full-text radiology reports, collected from 25,692 distinct CT studies conducted on 21,034 unique patients. Each CT volume is accompanied by a radiologist-authored report, which includes structured sections such as Impression, Findings, Patient Information, and Scan Technique as well as 18 pathology labels (see table 1 in Appendix).  Scans of 100 patients from the official training set were withheld for model selection. We used the official validation set of 3002 volumes from 1,564 studies to calculate the final metrics while filtering out all brain scans.
\subsubsection{RAD-CHEST}
RAD-ChestCT \cite{draelos2021machine} is a large-scale dataset comprising 36,000 chest CT scans from 20K unique patients. However, the current release includes an initial subset of 3,630 scans, representing roughly 10\% of the full dataset. Each scan is annotated with 84 abnormalities across 52 anatomical locations extracted from radiology reports. We selected 39 abnormality labels related to the lungs, opacity patterns, and the heart, as categorized in the RAD-Chest paper (see Table 1 in Appendix). We further categorized these 39 abnormality labels into 32 RAD-Chest-unique labels not present in the CT-RATE labels, and 7 shared labels that overlap with the CT-RATE labels for benchmarking.
\subsubsection{INSPECT}
INSPECT \cite{huang2023inspect} is a large-scale, multimodal dataset designed to enable research on integrating 3D CT pulmonary angiography (CTPA) with structured clinical data for pulmonary embolism (PE) patients. It includes de-identified longitudinal records from 19,438 patients, comprising 20,078 CT scans, 21,266 radiology report impression sections, and structured EHR data. In contrast to the other two datasets that are primarily diagnosis focused, INSPECT provides 1 diagnostic label (pulmonary embolism) and 7 prognostic labels (see table 1 in appendix). We note that the CT pulmonary angiography was performed with a contrast agent. At the time of writing, the information of the cohort hasn’t been made public. We thus split the dataset in 15148, 1000, 3000 scans for training, validation and testing respectively to be consistent with the setting in INSPECT \cite{huang2023inspect}. We dropped 145 scans that either had a corrupted affine matrix or dual channels. The percentage of positive labels of  our validation splits are consistent with those reported by INSPECT \cite{huang2023inspect}. 

In addition to these labels, we further generated 12 diagnosis labels as additional benchmarking (see table 1 in appendix). For these labels, we first applied keyword matching to locate a subset of reports that potentially include direct mention of the abnormality. We then randomly sampled equal numbers of reports as potential negatives. We then prompted gpt-4o-mini to validate the presence of the abnormality based on the impression and used these generated labels for validation (See the prompt in Appendix Table 2). A senior radiologist reviewed 120 samples, 10 per class randomly sampled, and found 115 (96\%) of those annotations were accurate. Since only the impression sections were available, we used word clouds for visualization and selected diagnosis labels that were frequently mentioned in the corpus. However, labels such as 'nodules' may include false negatives that cannot be identified based solely on the impression section.

\begin{table}[htbp]
\centering
\caption{Summary of Three Primary Datasets. See section 3.3 and appendix table 1 for definitions of shared and unique tasks.}
\resizebox{\columnwidth}{!}{%
\begin{tabular}{l ccc cc}
\toprule
 & \multicolumn{3}{c}{Number of CT Scans} & \multicolumn{2}{c}{Number of Tasks} \\
\cmidrule(lr){2-4} \cmidrule(lr){5-6}
 & Train & Validation & Test & Shared & Unique \\
\midrule
CT-RATE & 46168 & 221 & 3002 & 18 & -- \\
INSPECT & 15148 & 1000 & 3000 & 12 & 6 \\
RADCHEST & 2286 & 984 & 360 & 7 & 32 \\
\bottomrule
\end{tabular}%
}
\label{tab:dataset_summary}
\end{table}

\subsection{Imaging preprocessing}
All images were first converted to the NIfTI format when not originally downloaded in this format. Voxel intensities were standardized to Hounsfield Units (HU) and clipped to the range from -1000 to 1000 to suppress extreme outliers. Full-body segmentation was applied to each image scan using Total Segmentator, except for those from CT-RATE, for which we used the provided segmentation to ensure reproducibility. Images were cropped and centered using the following workflow: a thresholding operation was applied to the soft-tissue channel (HU -150 to 250) to identify the largest connected component. A convex hull transformation was then applied on each axial slice to estimate a common center of mass in the x–y plane. Image slices that are 5 mm away in the z axis from the top/bottom of the lung, identified by total Segmentator, were discarded. The voxel size was normalized to 1 mm x 1 mm x 1 mm. We cropped or padded when necessary to make images to  416 mm in x, 336 mm in y and  256 mm in z During training we applied augmentations including random rotation up to $\pm$ 15 degree, random zooming up to 10\% as well as adding Gaussian noise and random offset tn pixel intensity. CT volumes were then center cropped to a volume of 384 mm x 320 mm x 256 mm and resized to a 256 x 160 x 128 array for training and inference.

\subsection{Model training details}
We use EfficientNet b0 \cite{tan2019efficientnet} as the backbone for the vision encoder. The vision encoder was initialized with 2D model weights pretrained on ImageNet. The weights were inflated to adapt for 3D input by extending the 2D convolutional filters into the third dimension \cite{solovyev20223d}. Specifically, each 2D filter (e.g., a 3×3 kernel) was replicated along the depth axis to form a 3D kernel (e.g., a 3×3×3 filter), effectively copying the same weights across the third dimension. 

We used 4 stacked attention layers to integrate vision embeddings and text embeddings. We initialized the weights using the first 4 layers of PubMedBert \cite{gu2021domain}. AdamW was used as the optimizer with linearly decaying step sizes and 25 warm-up steps. As each dataset has different sample sizes, we iterated the model training for 25,000 steps with a batch size of 64 samples for most of our experiments while we keep track of validation metrics for every 5000 steps. For experiments that use only RADCHEST and INSPECT with significantly less vision-language task samples, we iterated for 10,000 steps with a batch size of 64 samples and keep track of validation metrics for every 1000 steps. Models were selected based on the best average Area under ROC curve (AUROC) across validation tasks. We trained our models on a single H100 GPU and most experiments were done in 24 hours.

\section{Experiments}
We trained models on various dataset configurations, including three standalone datasets CT-RATE, INSPECT and RADCHEST. We also trained models on combinations such as CT-RATE + INSPECT or CT-RATE + RADCHEST, and combined data that incorporated additional segmentation tasks. The notation CT-RATE + INSPECT denotes a combination of datasets, while +SEG indicates that segmentation tasks were included during training. All models were trained with 25,00 iterations with exception that INSPECT and RADCHEST alone were trained for 10,000 iterations. Table 1 summarizes the key characteristics of each dataset.

We then evaluated the classification accuracies on each dataset. For INSPECT and RADCHEST, we divided the tasks into two categories: those shared with CT-RATE as external validation to access out-of-distribution generalization, and those unique to each dataset for zero-shot validation. We evaluate the performance by the average area under ROC curve (AUROC) and the area under the precision-recall curve (AUPR) for each category.

For comparison, we implemented a baseline using a 3D EfficientNet-b0 backbone from the timm-3d library. The architecture includes depthwise separable convolutions, squeeze-and-excitation blocks, and global average pooling, followed by a classification head with two dropout layers and an intermediate ReLU activation. The model outputs a vector of logits, one per task, with each task trained independently using binary cross entrpoy. Optimization was performed using AdamW with a ReduceLROnPlateau learning rate scheduler. The same image augmentations were applied during training. Performance was evaluated using Area under the ROC curve (AUROC) and Area under the precision-recall curve (AUPR) using the same sample splits.

\section{Results}
Table 2 summarizes the average area under the ROC curve (AUROC) for our results. We also list the average area under the precision-recall curve in Appendix Table 3. We present our observations below based on AUROC, due to its robustness to the number of labels and the ease of cross-referencing performance reported in the literature. We include the AUROC for each individual category in Appendix Figures 1, 2, and 3.

\begin{table*}[t]
\centering
\caption{Average Area Under the ROC Curve (AUROC)}
\renewcommand{\arraystretch}{1.3}
\resizebox{\textwidth}{!}{%
\begin{tabular}{llc cc cc}
\hline
Model & Training Dataset & CTRATE & \multicolumn{2}{c}{INSPECT} & \multicolumn{2}{c}{RADCHEST} \\
\cline{4-7}
 & & & Shared & Unique & Shared & Unique \\
\hline
\multirow{8}{*}{Uniferum} & CTRATE & 0.8296 & 0.7207 & 0.5808 & 0.7960 & 0.5559 \\
 & CTRATE+SEG & 0.8311 & 0.7436 & 0.6130 & 0.8009 & 0.5768 \\
 & CTRATE+INSPECT & 0.8241 & 0.7186 & 0.7318 & 0.7933 & 0.5974 \\
 & CTRATE+INSPECT+SEG & 0.8298 & 0.6977 & 0.7383 & 0.8022 & 0.6077 \\
 & CTRATE+RADCHEST & 0.8286 & 0.7347 & 0.5929 & 0.8143 & 0.7417 \\
 & CTRATE+RADCHEST+SEG & 0.8305 & 0.7582 & 0.6124 & 0.8249 & 0.7512 \\
 & INSPECT & 0.6431 & 0.5584 & 0.7158 & 0.6097 & 0.5800 \\
 & RADCHEST & 0.7006 & 0.6379 & 0.4993 & 0.7668 & 0.6673 \\
\hline
CNN & INSPECT & -- & -- & 0.6132 & -- & -- \\
 & RADCHEST & -- & -- & -- & 0.700 & 0.634 \\
\hline
CT-NET & CTRATE & 0.6291\textsuperscript{1} & -- & -- & -- & -- \\
\hline
CT-CLIP (zero-shot) & CTRATE & 0.7311\textsuperscript{1} & -- & -- & -- & -- \\
\hline
CT-CLIP (vocabFine) & CTRATE & 0.7561\textsuperscript{1} & -- & -- & -- & -- \\
\hline
CT-CLIP (classFine) & CTRATE & 0.7561\textsuperscript{1} & -- & -- & -- & -- \\
\hline
CNN+Transformer & INSPECT & -- & -- & 0.6585\textsuperscript{2} & -- & -- \\
\hline
CNN & RADCHEST & -- & -- & -- & 0.8353\textsuperscript{3} & -- \\
\hline
\end{tabular}%
}
\label{tab:auroc_summary}
\\
{\raggedright \textsuperscript{$1$} \textup{Results from Hamamci et al.~\cite{hamamci2024developing}}\par}

{\raggedright \textsuperscript{$2$} \textup{Results from Huang et al. \cite{huang2023inspect} We note that the model was trained and evaluated on splits different from ours.}\par}
{\raggedright \textsuperscript{$3$} \textup{Results from Draelos et al. \cite{draelos2021machine} The model was trained on the data set 10 fold larger than publicly released.} \par}
\end{table*}

\subsection{Performance Comparison of Models Trained on a individual dataset}
We observe that Uniferum outperforms other models across multiple datasets. On the CT-RATE dataset, Uniferum achieved an average AUROC improvement of 7\% over CT-CLIP (83.0\% v.s 75.6\%) and 20\% over CT-NET, a convolutional neural networks model (83.0\% v.s. 62.9\%).

On the INSPECT dataset, Uniferum demonstrated a 10\% improvement over convolutional neural networks (71\% v.s. 61\%). Uniferum demonstrated 6\% improvement in auroc compared to the image only model reported in INSPECT \cite{huang2023inspect} (71\% v.s. 65.9\%), despite underperforming in diagnosing pulmonary embolism  (64.8\% v.s. 72.1\%). We note that, since the details of the INSPECT cohort have not been publicly released, we conducted evaluation using our own train-test split. Therefore, performance metrics may not be directly comparable to those reported in INSPECT \cite{huang2023inspect}. 

One the RADCHEST dataset, Uniferum outperformed our own CNN baseline by 6\% (76.7\% v.s. 70.0\% on RADCHESTshared, 66.7\% v.s 63.4\% on RADCHEST-unique). However, its performance was slightly lower than the results of the CNN model reported in RADCHEST \cite{draelos2021machine} (76.7\% v.s. 83.5\% on RADCHEST shared tasks). We emphasize that, because only 10\% of the RADCHEST samples were publicly available for training and evaluation, these comparisons should be interpreted with caution.

\subsection{Performance Comparison of Models Trained on Dataset Combinations}
We generally observed improved performance when incorporating additional datasets alongside CT-RATE. When comparing models trained on CT-RATE alone versus CT-RATE + INSPECT, performance remained comparable on tasks shared with CT-RATE. However, we observed a 4\% improvement on tasks unique to RADCHEST. A similar trend was observed when comparing CT-RATE to CT-RATE + RADCHEST: performance on CT-RATE tasks remained stable, while there was a 1\% improvement on classification tasks, both those shared with and unique to INSPECT.
We also found that including segmentation tasks consistently improved performance by 1–3\% compared to counterparts trained without segmentation. This held true for both shared-task validation and zero-shot scenarios. Notably, these segmentation tasks were derived from CT scans, suggesting that adding body segmentation represents a broadly applicable and low-cost method to enhance classification performance. Surprisingly, despite intentionally including the LUNA16 dataset that provides fine-grained lung nodule localization, we did not observe any performance gain in predicting the presence of lung nodules (see Appendix Table 4).

To demonstrate robustness, we observed similar patterns when replacing the pretrained attention layers with the first four layers of GatorTron-base \cite{yang2022large}, a model pretrained on clinical notes rather than scientific literature. Detailed results are reported in the appendix table 5.

\subsection{Evaluating Out-of-Distribution Generalization of Uniferum}
To assess cross-dataset generalizability, we evaluated the model on tasks common to multiple datasets. Specifically, we trained models exclusively on CT-RATE and assessed their performance using samples with the same labels from INSPECT and RADCHEST. The model achieved an AUROC of 83.0\% on CT-RATE, indicating strong out-of-distribution performance. When applied to related tasks in INSPECT, the AUROC dropped to 72.1\%, reflecting a moderate decline in generalization performance. On RADCHEST, the model achieved an AUROC of 79.6\%, suggesting better cross-dataset generalization than on INSPECT.

\subsection{Evaluating Zero-Shot Classification of Uniferum}
To assess the capability of zero-shot classification, we evaluated the model on tasks that are unique to each individual dataset. Specifically, we applied models trained on CT-RATE and assessed their performance on seven prognostic tasks and one pulmonary embolism classification task in INSPECT, as well as 32 nuanced classification labels in RADCHEST. Though we did not expect strong zero-shot performance, the model was able to achieve reasonable performance on certain zero-shot tasks (Figure 3 and 4) performing on par with or better than CNN baseline trained on in-distribution datasets. For example, the model trained on CT-RATE was able to predict 1-month, 6-month and 12 month in-hospital mortality with $71.6\pm 3.9\%$, $69.5 \pm 2.9\%$ and $68.2 \pm 2.8\%$ AUROC, as well as such unseen diagnostic tasks such as the presence of honeycombing pattern ($69.2 \pm 11.0\%$ AUROC, 10 positives out of 360 samples), coronary artery bypass grafting ($74.8 \pm 9.1\%$ AUROC, 16 positives out of 360 samples) and  pacemaker/defib ($90.5 \pm 7.7\%$ AUROC, 17 positives out of 360 samples) in RADCHEST. Such zero-shot capability could be raised by adding body segmentation tasks (Figure 3 and 4). For example, when trained on CTRATE+SEG dataset, the performance changes to $65.0\pm 13.\%$, $80.3 \pm 5.2\%$, $92.7\pm 5.2\%$ for predicting honeycombing pattern, coronary artery bypass grafting and pacemaker/defib, despite the improvement not statistically significant. Similarly, when evaluated on INSPECT, the performance improves to $75.6 \pm 3.8\%\; (p=0.20)$, $75.6 \pm 2.5\%\; (p = 0.03)$, $74.2 \pm 2.5\%\; (p=0.02)$ for 1, 6 and 12 month mortality respectively, with the inclusion of additional body segmentation tasks. The 95\% confidence intervals and p-values above were calculated using DeLong's method. We include the AUROC for each individual category in Appendix Figures 1, 2, and 3.

\begin{figure}
    \centering
    \includegraphics[width=\columnwidth]{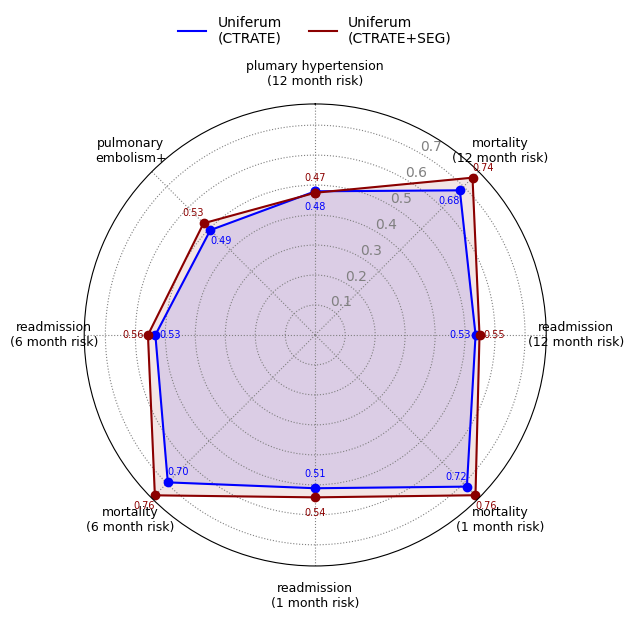}
    \caption{Zero-shot performance on INSPECT. Models were trained on CT-RATE with (red) and without (blue) body segmentation tasks and evaluated on tasks that are not included in CT-RATE.}
    \label{fig:fig3}
\end{figure}
\begin{figure}
    \centering
    \includegraphics[width=\columnwidth]{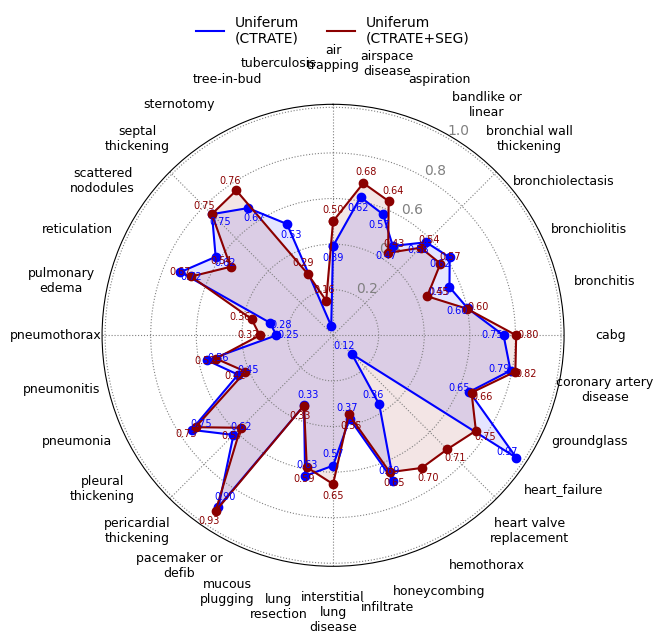}
    \caption{Zero-shot performance on RADCHEST. Models were trained on CT-RATE with (red) and without (blue) body segmentation tasks and evaluated on tasks that are not included in RADCHEST.}
    \label{fig:fig4}
\end{figure}

\section{Conclusion}
In this work, we present Uniferum, a volumetric vision-language model designed to overcome key limitations in medical imaging by unifying heterogeneous annotation types within a single training framework. By harmonizing multiple public 3D CT datasets and incorporating both classification and segmentation supervision, Uniferum demonstrates substantial gains in discriminative performance over existing models.

Our experiments show that Uniferum achieves 83\% in AUROC across 18 CT-RATE abnormality predictions, outperforming both CLIP-based baselines and standard multi-label CNNs. Additionally, Uniferum maintains strong generalization ability, achieving 72\% and 79\% AUROC on out-of-distribution datasets RAD-CHEST and INSPECT respectively. Notably, we observe that adding body segmentation tasks contributes to 1-3\% performance gains, highlighting the value of integrating diverse supervision signals.

These findings underscore the importance of flexible training strategies in high-stakes domains like radiology. By enabling robust and data-efficient learning from fragmented and variably annotated datasets, Uniferum represents a step toward clinically viable vision-language models for 3D CT interpretation. Future work will focus on scaling to additional modalities, enhancing cross-task synthesis and validating performance in real-world clinical workflows. For those about to fight for VLM, diverse supervision signals abound in public volumetric datasets.

{\noindent\bf{Acknowledgment}}
This work was supported by Scientific Computing and Data at the Icahn School of Medicine at Mount Sinai and NIH grants UL1TR004419, S10OD026880, and S10OD030463. XM is supported by the Eric and Wendy Schmidt AI in Human Health Fellowship, a program of Schmidt Sciences. This research used data provided by the Stanford Center for Artificial Intelligence in Medicine and Imaging (AIMI).
{
    \small
    %\clearpage
    \bibliographystyle{ieeenat_fullname}

    \bibliography{main}
}
\clearpage
\setcounter{page}{1}
\onecolumn
\maketitlesupplementary

\setcounter{table}{0}
\setcounter{figure}{0}

\vspace{\baselineskip}
\noindent \textbf{}

%%%%%%%%
\begin{table*}[ht]
\caption{List of tasks for training and validation.}
\begin{tabular}{|l|p{0.75\textwidth}|}
\hline
CTRATE & Arterial wall calcification, Atelectasis, Bronchiectasis, Cardiomegaly, Consolidation, Coronary artery wall calcification, Emphysema, Hiatal hernia, Interlobular septal thickening, Lung nodule, Lung opacity, Lymphadenopathy, Medical material, Mosaic attenuation pattern, Peribronchial thickening, Pericardial effusion, Pleural effusion, Pulmonary fibrotic sequela \\ \hline
INSPECT (shared) & 12\_month\_PH, 12\_month\_mortality, 12\_month\_readmission, 1\_month\_mortality, 1\_month\_readmission, 6\_month\_mortality, 6\_month\_readmission, pe\_positive \\ \hline
INSPECT (unique) & Atelectasis, Bronchiectasis, Cardiomegaly, Consolidation, Coronary artery wall calcification, Emphysema, Hiatal hernia, Interlobular septal thickening, Lung nodule, Peribronchial thickening, Pericardial effusion, Pleural effusion \\ \hline
RADCHEST (shared) & Atelectasis, Bronchiectasis, Cardiomegaly, Consolidation, Emphysema, Pericardial effusion, Pleural effusion \\ \hline
RADCHEST (unique) & Air trapping, Airspace disease, Aspiration, Bandlike or linear, Bronchial wall thickening, Bronchiolectasis, Bronchiolitis, Bronchitis, CABG, Coronary artery disease, Ground-glass, Heart failure, Heart valve replacement, Hemothorax, Honeycombing, Infiltrate, Interstitial lung disease, Lung resection, Mucous plugging, Pacemaker or defib, Pericardial thickening, Pleural thickening, Pneumonia, Pneumonitis, Pneumothorax, Pulmonary edema, Reticulation, Scattered nodules, Septal thickening, Sternotomy, Tree-in-bud, Tuberculosis \\ \hline
Body segmentation tasks & Liver, Stomach, Left upper pulmonary lobe, Left lower pulmonary lobe, Right upper pulmonary lobe, Right middle pulmonary lobe, Right lower pulmonary lobe, Esophagus, Tracheal airway, Thoracic vertebra 1-12, Cardiac organ, Thoracic aorta, Pulmonary vein, Left atrial appendage, Superior vena cava, Inferior vena cava, Left rib 1-12, Right rib 1-12, Sternal body, Costal cartilaginous structures, Lungs, Thoracic vertebrae, Ribs, Right ribs, Left ribs \\ \hline
\end{tabular}
\end{table*}

%%%%%%%%
\begin{table*}[t]
\centering
\caption{Prompt for identifying presence of abnormality in radiology reports.}
\begin{tabular}{|p{0.95\textwidth}|}
\hline
Please review the radiology report and identify any findings indicative of \{PATHOLOGY\}, focusing on the lung region and heart. Return the following in a structured JSON format:\\[1ex]
\{\{\\
\quad "KeySentence": "Provide the key sentence in the report that supports your answer. If no relevant sentences, return 'NONE'.",\\
\quad "PresenceOfPathology": "Answer with 'Yes' or 'No' based on the presence of \{PATHOLOGY\}. If no pathology is mentioned, assume 'No'."\\
\}\}\\[1ex]
Return nothing else.\\[1ex]
Below is the report:\\
\{REPORT\} \\
\hline
\end{tabular}
\end{table*}

\begin{table*}[t]
\caption{Performance comparison on classifying Lung nodules}
\begin{tabular}{|l|c|c|c|}
\hline
\multicolumn{1}{|r|}{model} & training dataset    & val dataset           & auroc  \\ \hline
\multirow{8}{*}{Uniferum}   & CTRATE              & CTRATE (Lung nodules) & 0.6915 \\ \cline{2-4} 
                            & CTRATE+SEG          & CTRATE (Lung nodules) & 0.6791 \\ \cline{2-4} 
                            & CTRATE+INSPECT      & CTRATE (Lung nodules) & 0.6838 \\ \cline{2-4} 
                            & CTRATE+INSPECT+SEG  & CTRATE (Lung nodules) & 0.68   \\ \cline{2-4} 
                            & CTRATE+RADCHEST     & CTRATE (Lung nodules) & 0.6873 \\ \cline{2-4} 
                            & CTRATE+RADCHEST+SEG & CTRATE (Lung nodules) & 0.6884 \\ \cline{2-4} 
                            & INSPECT             & CTRATE (Lung nodules) & 0.4891 \\ \cline{2-4} 
                            & RADCHEST            & CTRATE (Lung nodules) & 0.463  \\ \hline
\end{tabular}
\end{table*}

\begin{table*}[t]
\caption{Average AUROCs of Uniferum initialized by GatorTron}
\begin{tabular}{|l|ccccc|}
\hline
\multicolumn{1}{|c|}{training dataset} & \multicolumn{5}{c|}{validation}                                                                                                \\ \hline
                                       & \multicolumn{1}{c|}{CTRATE} & \multicolumn{2}{c|}{INSPECT}                              & \multicolumn{2}{c|}{RADCHEST}        \\ \hline
                                       & \multicolumn{1}{l|}{}       & \multicolumn{1}{c|}{shared} & \multicolumn{1}{c|}{unique} & \multicolumn{1}{c|}{shared} & unique \\ \hline
CTRATE                                 & \multicolumn{1}{c|}{0.8255} & \multicolumn{1}{c|}{0.7505} & \multicolumn{1}{c|}{0.5872} & \multicolumn{1}{c|}{0.7933} & 0.539  \\ \hline
CTRATE+INSPECT                         & \multicolumn{1}{c|}{0.829}  & \multicolumn{1}{c|}{0.6896} & \multicolumn{1}{c|}{0.735}  & \multicolumn{1}{c|}{0.79}   & 0.5896 \\ \hline
CTRATE+INSPECT+SEG                     & \multicolumn{1}{c|}{0.8291} & \multicolumn{1}{c|}{0.7302} & \multicolumn{1}{c|}{0.7412} & \multicolumn{1}{c|}{0.7904} & 0.6174 \\ \hline
CTRATE+RADCHEST                        & \multicolumn{1}{c|}{0.8278} & \multicolumn{1}{c|}{0.7432} & \multicolumn{1}{c|}{0.5714} & \multicolumn{1}{c|}{0.8234} & 0.748  \\ \hline
CTRATE+RADCHEST+SEG                    & \multicolumn{1}{c|}{0.8321} & \multicolumn{1}{c|}{0.7692} & \multicolumn{1}{c|}{0.5777} & \multicolumn{1}{c|}{0.8205} & 0.7298 \\ \hline
CTRATE+SEG                             & \multicolumn{1}{c|}{0.8305} & \multicolumn{1}{c|}{0.746}  & \multicolumn{1}{c|}{0.5867} & \multicolumn{1}{c|}{0.785}  & 0.5589 \\ \hline
\end{tabular}
\end{table*}

%%%%%%%%%%%%
\begin{table*}[t]
\caption{Area under precision recall curve. Model initialized by PubMedBert}
\begin{tabular}{|l|l|ccccc|}
\hline
\multicolumn{1}{|c|}{model} & \multicolumn{1}{c|}{training dataset} & \multicolumn{5}{c|}{validation}                                                                                                \\ \hline
                            &                                       & \multicolumn{1}{c|}{CTRATE} & \multicolumn{2}{c|}{INSPECT}                              & \multicolumn{2}{c|}{RADCHEST}        \\ \hline
                            &                                       & \multicolumn{1}{l|}{}       & \multicolumn{1}{c|}{shared} & \multicolumn{1}{c|}{unique} & \multicolumn{1}{c|}{shared} & unique \\ \hline
\multirow{8}{*}{Uniferum}   & CTRATE                                & \multicolumn{1}{c|}{0.5583} & \multicolumn{1}{c|}{0.7317} & \multicolumn{1}{c|}{0.1677} & \multicolumn{1}{c|}{0.5274} & 0.1617 \\ \cline{2-7} 
                            & CTRATE+INSPECT                        & \multicolumn{1}{c|}{0.5377} & \multicolumn{1}{c|}{0.7247} & \multicolumn{1}{c|}{0.2949} & \multicolumn{1}{c|}{0.5171} & 0.1565 \\ \cline{2-7} 
                            & CTRATE+INSPECT+SEG                    & \multicolumn{1}{c|}{0.5543} & \multicolumn{1}{c|}{0.7221} & \multicolumn{1}{c|}{0.2998} & \multicolumn{1}{c|}{0.5486} & 0.165  \\ \cline{2-7} 
                            & CTRATE+RADCHEST                       & \multicolumn{1}{c|}{0.5505} & \multicolumn{1}{c|}{0.747}  & \multicolumn{1}{c|}{0.172}  & \multicolumn{1}{c|}{0.5701} & 0.2566 \\ \cline{2-7} 
                            & CTRATE+RADCHEST+SEG                   & \multicolumn{1}{c|}{0.5565} & \multicolumn{1}{c|}{0.7671} & \multicolumn{1}{c|}{0.1703} & \multicolumn{1}{c|}{0.5862} & 0.2452 \\ \cline{2-7} 
                            & CTRATE+SEG                            & \multicolumn{1}{c|}{0.5557} & \multicolumn{1}{c|}{0.7555} & \multicolumn{1}{c|}{0.18}   & \multicolumn{1}{c|}{0.5408} & 0.1476 \\ \cline{2-7} 
                            & INSPECT                               & \multicolumn{1}{c|}{0.2859} & \multicolumn{1}{c|}{0.6023} & \multicolumn{1}{c|}{0.2661} & \multicolumn{1}{c|}{0.2552} & 0.109  \\ \cline{2-7} 
                            & RADCHEST                              & \multicolumn{1}{c|}{0.3588} & \multicolumn{1}{c|}{0.668}  & \multicolumn{1}{c|}{0.1259} & \multicolumn{1}{c|}{0.4587} & 0.1848 \\ \hline
\end{tabular}
\end{table*}
%%%%%%%%%%%
\newpage

\begin{figure*}[ht]
    \centering
    \includegraphics[width=250pt]{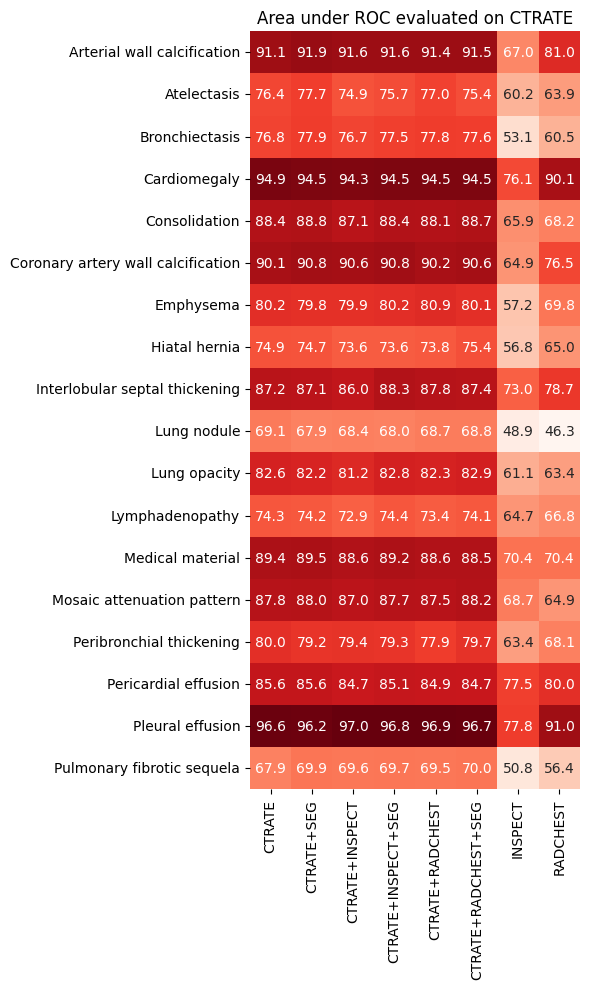}
    \caption{AUROC heatmap on CTRATE. The x-axis represents different training datasets, while the y-axis corresponds to the target classes. Color intensity indicates the area under the ROC curve (AUROC), with higher values reflecting better classification performance.}
    \label{fig:app_fig1}
\end{figure*}
%%%%%%%%%%
%%%%%%%%%%
\begin{figure*}[ht]
    \centering
    \includegraphics[width=250pt]{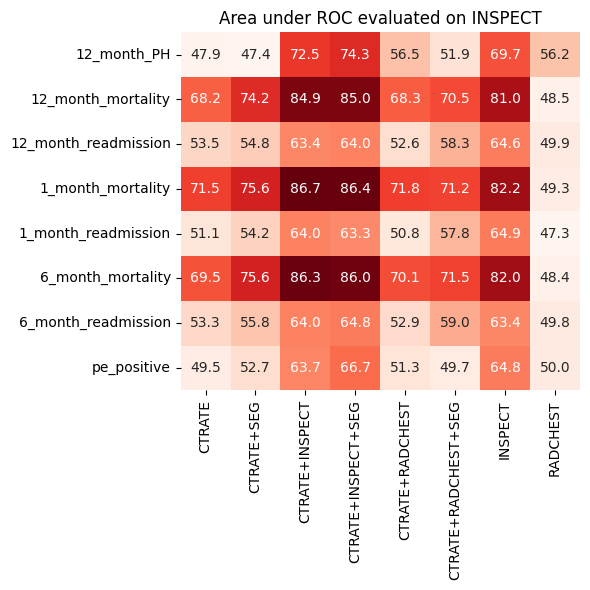}
    \caption{AUROC heatmap on INSPECT. The x-axis represents different training datasets, while the y-axis corresponds to the target classes. Color intensity indicates the area under the ROC curve (AUROC), with higher values reflecting better classification performance.}
    \label{fig:app_fig2}
\end{figure*}
%%%%%%%%%%
%%%%%%%%%%

\begin{figure*}[ht]
    \centering
    \includegraphics[width=0.45\textwidth, keepaspectratio]{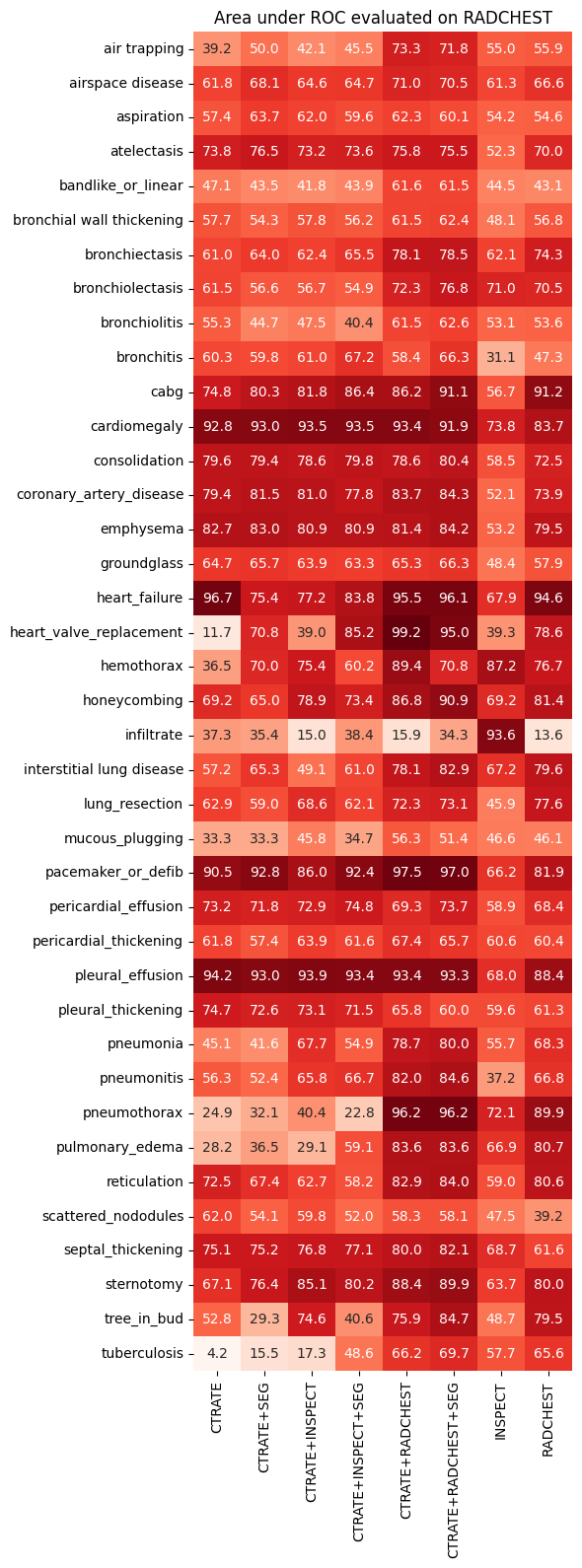}
    \caption{AUROC heatmap on RADCHEST. The x-axis represents different training datasets, while the y-axis corresponds to the target classes. Color intensity indicates the area under the ROC curve (AUROC), with higher values reflecting better classification performance.}
    \label{fig:app_fig3}
\end{figure*}
%%%%%%%%%%
\end{document}